\documentclass{article}
\usepackage[a4paper]{geometry}
\usepackage{booktabs}
\usepackage{graphicx}
\usepackage{siunitx}
\usepackage{tabularx}
\usepackage{placeins}
\usepackage{amsmath}
\usepackage{amssymb}
\usepackage{float}
\usepackage{multirow}
\usepackage{tabularx}
\usepackage[version=4]{mhchem}
\usepackage{subcaption}
\usepackage{hyperref}
\usepackage{titling}
\usepackage{authblk}
\usepackage{array}
\usepackage{tikz}
\usepackage[inline]{enumitem}
\usepackage{yhmath}
\usepackage[version=4]{mhchem}
\usepackage{upgreek,textgreek}
\usepackage{subcaption}
\usepackage{comment}
\usepackage[
    natbib=true,
    style=numeric,
    sorting=none
]{biblatex}
\providecommand{\keywords}[1]
{
  \small	
  \textbf{\textit{Keywords---}} #1
}

\usepackage{xargs}                      % Use more than one optional parameter in a new commands
\usepackage[colorinlistoftodos,prependcaption,textsize=tiny]{todonotes}
\newcommandx{\unsure}[2][1=]{\todo[linecolor=red,backgroundcolor=red!25,bordercolor=red,#1]{#2}}
\newcommandx{\change}[2][1=]{\todo[linecolor=blue,backgroundcolor=blue!25,bordercolor=blue,#1]{#2}}
\newcommandx{\info}[2][1=]{\todo[linecolor=green,backgroundcolor=green!25,bordercolor=green,#1]{#2}}
\newcommandx{\improvement}[2][1=]{\todo[linecolor=Plum,backgroundcolor=Plum!25,bordercolor=Plum,#1]{#2}}
\newcommandx{\thiswillnotshow}[2][1=]{\todo[disable,#1]{#2}}
\usetikzlibrary{fadings}
\usetikzlibrary{patterns}
\usetikzlibrary{shadows.blur}
\usetikzlibrary{shapes}
\usepackage{lineno}
\usepackage{setspace}
\DeclareUnicodeCharacter{2212}{\textendash}
\DeclareUnicodeCharacter{2032}{\textendash}

\title{Symbolic regression for defect interactions in 2D materials}

\author[1]{Mikhail Lazarev\thanks{Corresponding author: mvlazarev@hse.ru}}

\author[3,2]{Andrey Ustyuzhanin}

\affil[1]{HSE University, Myasnitskaya Ulitsa, 20, Moscow, Russia, 101000}
\affil[2]{Institute for Functional Intelligent Materials, National University of Singapore, 4 Science Drive 2, Singapore 117544}
\affil[3]{Constructor University, Campus Ring 1, Bremen, 28759, Germany}

\makeatletter
\let\Title\@title
\makeatother
\usepackage[T2A]{fontenc}
\usepackage[english]{babel}
\usepackage{csquotes}

\begin{document}
\maketitle

\begin{abstract}
Machine learning models have become firmly established across all scientific fields. Extracting features from data and making inferences based on them with neural network models often yields high accuracy; however, this approach has several drawbacks. Symbolic regression is a powerful technique for discovering analytical equations that describe data, providing interpretable and generalizable models capable of predicting unseen data. Symbolic regression methods have gained new momentum with the advancement of neural network technologies and offer several advantages, the main one being the interpretability of results. In this work, we examined the application of the deep symbolic regression algorithm SEGVAE to determine the properties of two-dimensional materials with defects. Comparing the results with state-of-the-art graph neural network-based methods shows comparable or, in some cases, even identical outcomes. We also discuss the applicability of this class of methods in natural sciences.

\keywords{2D materials, Machine Learning, Symbolic Regression, GNNs, Interpretability.   \\[5pt]}

\end{abstract}

 %%\tableofcontents

\newpage{}

 %%\tableofcontents

 %%\flushbottom{}
 %%\maketitle{}
 %%\thispagestyle{empty}

 %%\addcontentsline{toc}{section}{Unnumbered Section}

\section{Introduction}
\begin{figure}[H]
	\noindent
	\centering
	\includegraphics[width=12.6cm]{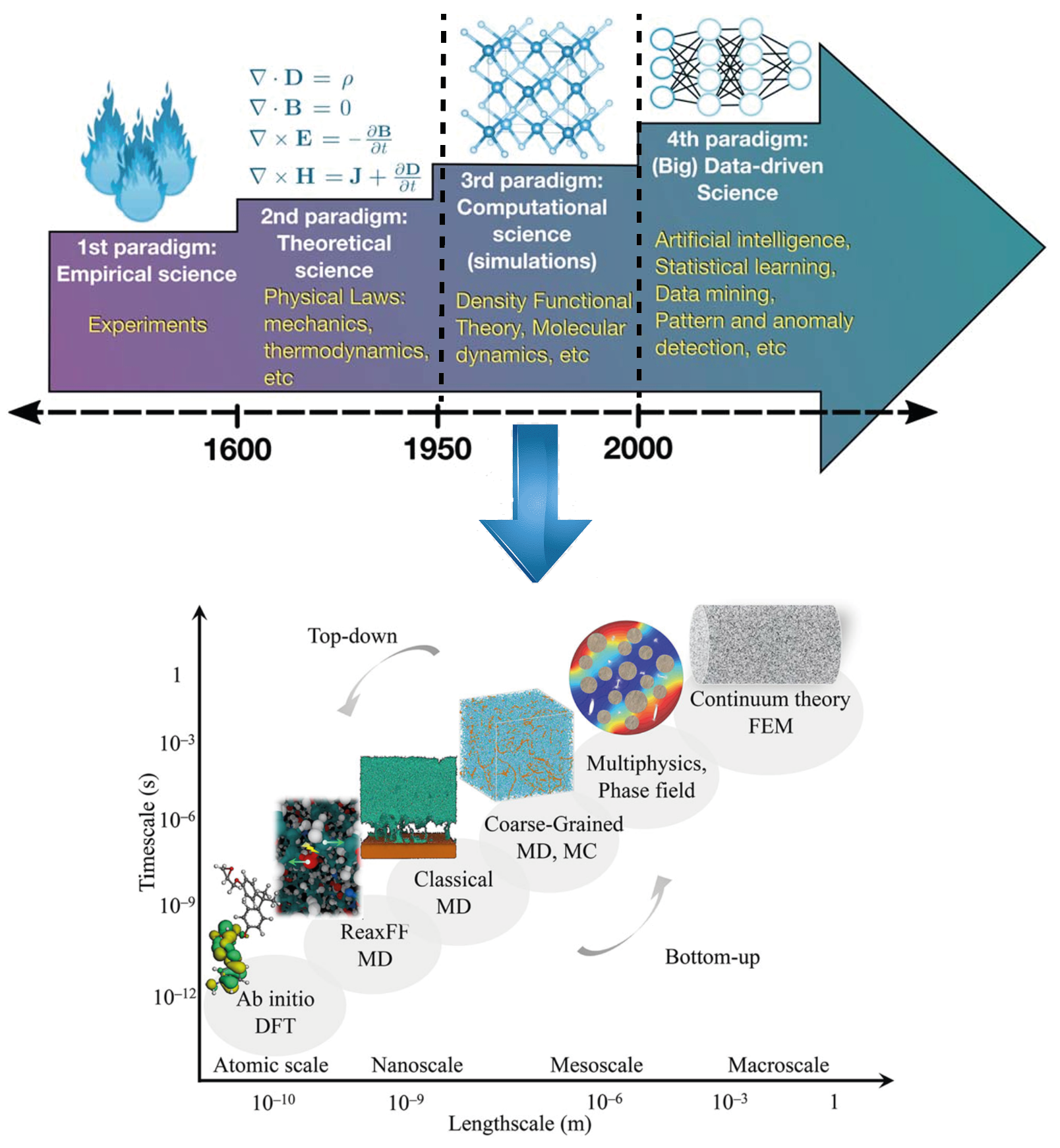}
	\caption{Upper part: Science paradigm over time. Taken from \cite{schleder2019dft}. Lower part: Simulation methods based on physics theories. Taken from \cite{lin2023multiscale}.}
	\label{fig:Intro}
\end{figure}

Since ancient times, humanity has sought to understand dependencies in observable effects and phenomena around it. The new discoveries that have driven us to the technological advancements of today were made manually through an iterative process of theory building and practical validation. These theories, typically expressed in the language of mathematics, can be represented through formulas or symbols; for instance, the theory of classical electrodynamics is fully described by four equations. This analogy is not unique, as other fields or branches of science also have compact symbolic representations to describe specific effects. Abstractly, the development of scientific knowledge methods can be depicted in 4 paradigms as shown in Figure \ref{fig:Intro} (top section). The lower part of Figure \ref{fig:Intro} illustrates the types of numerical modeling methods for physical systems across different time and spatial scales.

Today, neural network-based technologies have permeated virtually all areas of life and science, reaching new heights in process automation and the search for dependencies in observed phenomena. On one hand, modern hardware enables the training of extremely large neural network algorithms, which have become the gold standard of quality in many fields. However, they function as black boxes, producing results that are often impossible to interpret. In science, interpretability is essential due to the need to uncover complex relationships within data. Thus, for neural network methods to be widely applicable, we must either develop algorithms to interpret them or design inherently interpretable algorithms. The interpretability of model predictions is crucial for many areas of potential application, such as medicine, autonomous vehicles, and finance—fields where errors could have significant consequences. A promising solution lies in symbolic regression methods, which integrate neural network approaches under the hood but yield interpretable final results.

Symbolic regression is a form of regression that searches for formulas in the space of mathematical expressions. The goal is to find an optimal expression, one that best explains the training data, while often requiring that the formula be as concise as possible to ensure interpretability. Such an interpretable model offers several advantages over black-box models; assuming the discovered formula accurately captures the system's underlying laws, it can be extrapolated beyond the training data.

The primary objective of symbolic regression is to uncover mathematical dependencies that describe observed data. While the topic of symbolic regression is not new, it predates the widespread use of neural networks. Historically, evolutionary programming methods such as genetic algorithms \cite{searson2010gptips} were used to generate formulas that best fit the data. There even existed commercial products, like Eureqa \cite{schmidt2009distilling}, which applied symbolic regression algorithms and were state-of-the-art at the time. However, today, a diverse range of neural network-based symbolic regression models has emerged, outperforming previous evolutionary algorithm-based approaches. The variety of symbolic regression algorithms is reviewed in appendices \ref{Appendix}.

\section{Related Work}

In natural science, symbolic regression has emerged as a powerful method for uncovering analytical relationships within complex datasets, providing interpretable models that align closely with underlying physical laws. In material science, symbolic regression (SR) is implemented for some tasks as well. For example, in \cite{wang2019symbolic} can be found a comprehensive review of the application of SR in materials science can be found, where a big part of examples has genetic algorithms (GA) under the hood.
Traditionally, GAs are still popular for SR tasks, for example, in \cite{sofos2022combined}, authors show that generated equations perform better in this case than well-established empirical equations for Lennard–Jones fluid descriptors. This is another evidence of SR application in the same field \cite{alam2022symbolic}. Genetic programming (GP) remains competitive for certain physics modeling tasks and can be combined with graph-based inductive biases; for example in \cite{reuter2023graph} demonstrate GP-based symbolic models for particle-laden flows and compare them against graph-network baselines. Discovering interatomic potentials from data with SR based on GA was demonstrated in \cite{burlacu2022symbolic}. Another use case where SR was implemented to establish a symbolic dependence form of a band gap in NaCl-type compounds \cite{wang2022symbolic}. In \cite{loftis2020lattice} GA GA-based SR was compared in terms of accuracy with other machine learning (ML) methods and demonstrated good performance; however, the final equation forms usually contain inner functions that usually do not appear in physical models.
An interesting study of atomic-scale modeling with SR for 2D materials \cite{xie2022machine} in this particular case, low-dimensional \ce{TiO_2}. Authors use SR to study automatically designed molecular descriptors that capture the essential chemical interactions, such as those influenced by external electric fields. Combined with machine learning, specifically random forest algorithms, symbolic regression achieves high accuracy in predicting adsorption energies. SR for searching new descriptors also was used in \cite{weng2020simple}. In \cite{hernandez2019fast} SR has been used to discover fast, accurate, and transferable many-body potentials (e.g., for Cu), reducing simulation cost while retaining interpretability; such closed forms enable larger design-space exploration and quick what-if analyses in process and device modeling. In \cite{zhang2024integrative} integrative SR pipelines for multicomponent perovskite oxides provide interpretable formulas used to screen large compositional spaces for stable phases before costly Density functional theory (DFT) or experiments. In the paper \cite{flores2022learning} authors demonstrated another approach to utilize SR, instead of use of an operator's library like in GA-SR, where they use a feature generation algorithm and select features based on multiple criteria, thus selecting a few out of thousands of features, and use them to present finale equation in a compact form akin to a linear function. 

Many other examples can be found in the literature where symbolic formulas are needed for some purposes like in \cite{kabliman2021application}, where the goal was to find a formula that can be integrated into the simulation tool. These examples illustrate how closed-form, human-readable equations function as “design rules” that can be embedded into screening loops, surrogate models, and process windows, complementing black-box ML.

%\cite{zheng2021symbolic}  Symbolic transformer accelerating machine learning screening

%\cite{reinbold2021robust}  learning from noisy, incomplete, highdimensional experimental data via physically constrained symbolic regression

\begin{figure}[H]
    \noindent
    \centering
    \includegraphics[width=\textwidth]{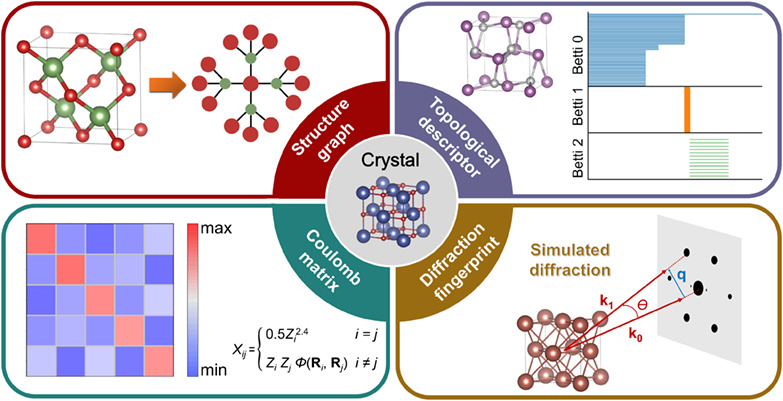}
    \caption{One crystal can be represented in various ways using different representations. These representations include: a graph containing atom and bond weights, Coulomb matrix,   diffraction fingerprint, using topological descriptors and etc. Taken from \cite{li2022encoding} with permission.}
    \label{fig:representations}
\end{figure}

Most of the symbolic regression methods discussed above rely on genetic algorithms (GA) for formula construction. As in machine learning models, selecting the correct descriptors is crucial for symbolic regression algorithms. A large feature space or an extensive set of arguments for the target formula results in a vast search space, which often leads to either a loss of interpretability to achieve the best quality metric or unsatisfactory outcomes. Every in silico object under study must be represented in a machine-readable format, and substantial datasets must be available for training machine learning models.

\begin{figure}[H]
    \noindent
    \centering
    \includegraphics[width=15.6cm]{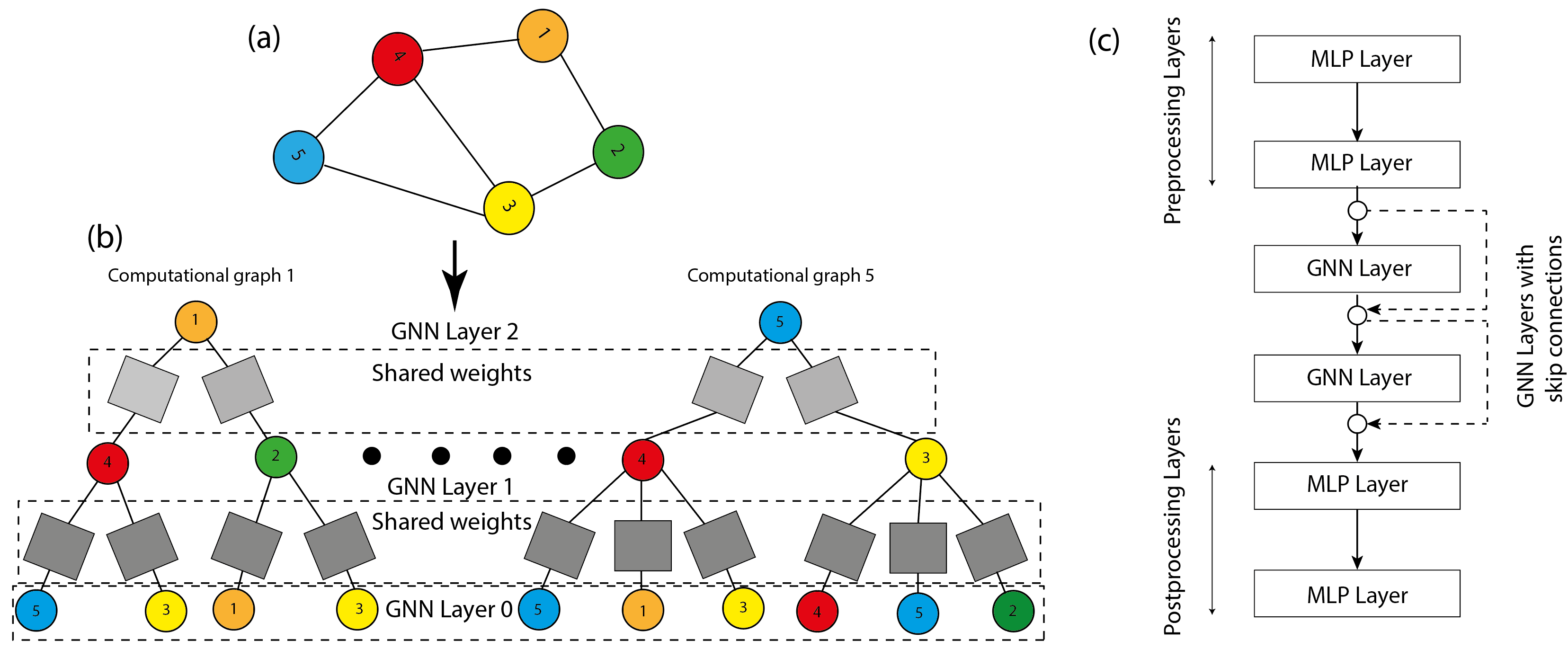}
    \caption{Convolutional GNN (a) Example of graph structure (b) For each node, a computational graph is constructed, where each layer contains aggregation functions with shared weights across that layer (c) Upscaled vision of сonvolutional GNN.}
    \label{fig:gnn}
\end{figure}

A crystal can be represented in various ways, some of which are illustrated in Figure \ref{fig:representations}. A graph representation of a crystal is intuitive and arguably the most common approach, with several neural algorithms in the form of graph neural networks (GNNs) specifically designed for crystal structures in this format \cite{zhang2023artificial}. It is essential to account for the periodic nature of the crystal lattice when constructing the graph. Graph neural networks (GNNs) operate on graphs by iteratively updating learnable embeddings of nodes (and optionally edges and global attributes) via message passing. At each layer, a node collects messages from its neighbors using an aggregation operator (e.g., sum/mean/attention) and combines them with its current state to produce an updated representation; after several layers, a readout/pooling operation produces graph-level predictions. In atomistic and crystal-property prediction, nodes typically encode atomic descriptors (e.g., element type/embedding, valence-related or periodic-table features), edges encode pairwise geometric information (e.g., interatomic distance expanded in a radial basis, bond type, neighbor shell), and a global state may store system-level metadata (e.g., charge, pressure/temperature, or composition statistics). In abstract terms, a GNN can be represented as shown in Figure \ref{fig:gnn}. 

\begin{figure}[H]
    \noindent
    \centering
    \includegraphics[width=12.6cm]{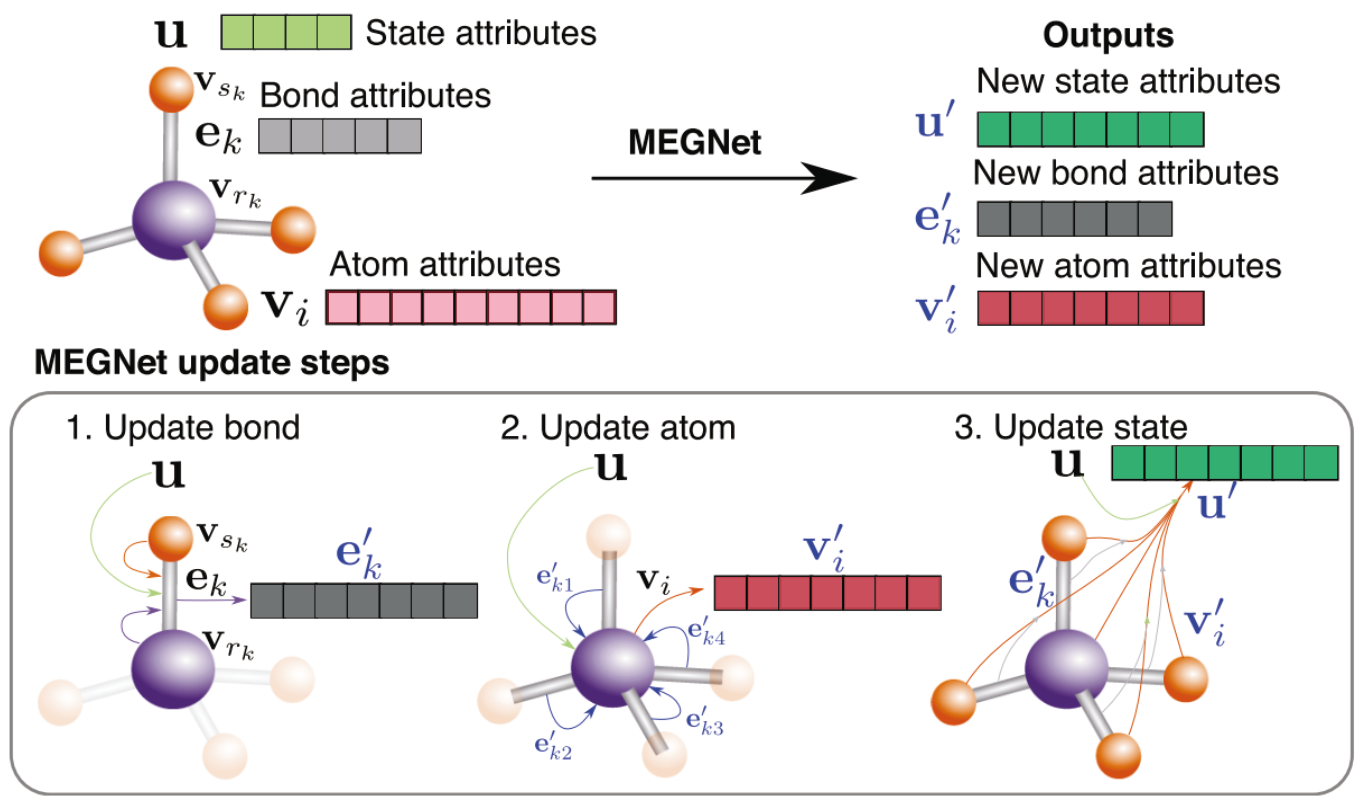}
    \caption{An overview of a MEGNet module. The initial graph is represented by the set of atomic attributes $V=v_i$, bond attributes $E =\{(e_k, r_k, s_k)\}$, and global state attributes $u$. In the first update step, the bond attributes are updated. Information flows from atoms that form the bond, the state attributes, and the previous bond attribute to the new bond attributes. Similarly, the second and third steps update the atomic and global state attributes, respectively, by information flow among all three attributes. The final result is a new graph representation. Taken from \cite{chen2019graph}.}
    \label{fig:megnet}
    
\end{figure}

One of the most powerful graph neural network (GNN) architectures for crystals is MEGNet family \cite{chen2019graph}, which has also performed well on this dataset, making it a suitable choice for training on sparse representations. A schematic of the MEGNet architecture is shown in Figure \ref{fig:megnet}, with its defining feature being the update mechanism for the vector representations of graph nodes, edges, and a global representation vector. However, GNNs suffer from a lack of interpretability or face overfitting challenges on small datasets, and most of the neural network (NN) approaches struggle to generalize to data outside the training distribution. In \cite{omee2024structure}, reported gains can reflect overfitting: an out-of-distribution (OOD) benchmark showed that random splits artificially inflate performance due to dataset redundancy and that state-of-the-art GNNs generalize poorly to unseen chemistries/structures.

%(раписать про эффективноть это архитектуры по стравнению с другими?)

Today, a wide range of crystal databases are accessible \cite{rashid2024review}, but due to the diversity of crystal types and target properties, it is often the case that few or no structures are available for specific tasks in open databases—particularly with experimental data. Given the cost and complexity of obtaining experimental data, most databases predominantly contain results from physical simulations. In our example, we also use a synthetic dataset, specifically a new dataset of two-dimensional materials with two types of defects: substitutions and vacancies. This dataset was computed using density functional theory (DFT) \cite{huang2023unveiling}. The types and details of the vacancies are listed in Table \ref{tab:defect-definitions}. 

\begin{table}[H]
    \centering
    \begin{tabular}{c|c|c}
         Material & Substitutions & Vacancies \\
         \ce{MoS_2} & \ce{S} $\rightarrow$ \ce{Se}; \ce{Mo} $\rightarrow$ \ce{W} & \ce{Mo}; \ce{S} \\
         \ce{WSe_2} & \ce{Se} $\rightarrow$ \ce{S}; \ce{W} $\rightarrow$ \ce{Mo} & \ce{W}; \ce{Se} \\
         \ce{h-BN} & \ce{B} $\rightarrow$ \ce{C}; \ce{N} $\rightarrow$ \ce{C} & \ce{B}; \ce{N} \\
         \ce{GaSe} & \ce{Ga} $\rightarrow$ \ce{In}; \ce{Se} $\rightarrow$ \ce{S} & \ce{Ga}; \ce{Se} \\
         \ce{InSe} & \ce{In} $\rightarrow$ \ce{Ga}; \ce{Se} $\rightarrow$ \ce{S} & \ce{In}; \ce{Se} \\
         BP & \ce{P} $\rightarrow$ \ce{N} & \ce{P}
    \end{tabular}
    \caption{Point defect types present in the datasets \cite{huang2023unveiling}}
    \label{tab:defect-definitions}
\end{table}

\begin{figure}[H]
	\noindent
	\centering
	\includegraphics[width=15.6cm]{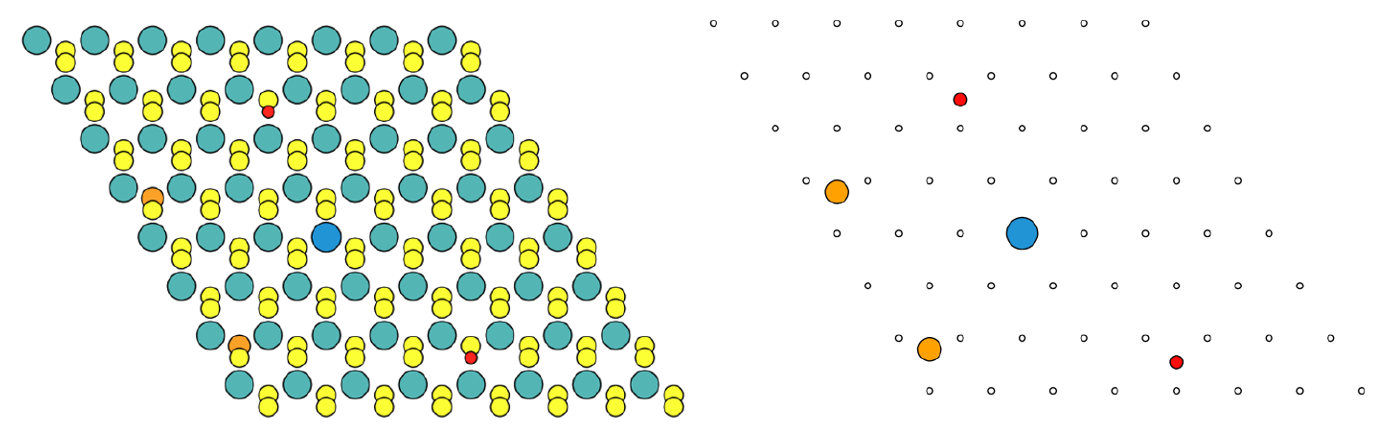}
	\caption{ Left side: pristine \ce{MoS2} crystal. Right side: it's a sparse representation (defects only). }
	\label{fig:Sparce_representations}
\end{figure}

It has been demonstrated that a neural network with this architecture predicts crystal properties more accurately than others available at the time. However, like other neural network approaches, it lacks interpretability and generalizability. For studying defects in crystals, a sparse representation—constructing a graph of defects as shown in Figure \ref{fig:Sparce_representations}—proved more effective. As shown in \cite{kazeev2023sparse}, this approach significantly improves model accuracy for predicting formation energy, although it does not yield similar improvements for predicting the highest occupied–lowest unoccupied molecular orbital (HOMO-LUMO) gap.

Neural networks often function as a black box; however, scientific inquiry demands a clear and precise understanding of observable physical parameters.  Here we want to fill the gap and recover physically constrained expressions for 2D defect properties from a small dataset and demonstrate its generalizability. We therefore seek interpretable, closed-form expressions that (i) encode physically plausible interaction kernels, (ii) generalize across different kinds of structures, and (iii) remain reproducible. The next section formalizes this task and introduces our SEGVAE-based method tailored to these requirements.

%\subsection{Materials and available data}

%\section{Property prediction}

\section{SEGVAE for 2D materials}

Uncovering the laws governing atomic interactions within a crystal lattice will not only enable the prediction of crystal properties in an interpretable form but will also pave the way for designing crystals with tailored properties.

We require an SR method that (i) explicitly encodes physical priors, (ii) controls expression complexity, and (iii) remains reproducible under small, noisy DFT datasets. Among genetic programming GP SR, deep/RL SR (e.g., Deep Symbolic Optimization (DSO)), and LLM-driven SR, we adopt SEGVAE \cite{popov2023symbolic} because it combines predicate-guided hypothesis restriction, structure pre-training, a Bank of Best Formulas, and Pareto selection over accuracy vs. complexity. Our method can be viewed within the estimation-of-distribution algorithm (EDA) paradigm for genetic programming: instead of relying solely on hand-crafted variation operators, we iteratively fit a probabilistic model to high-performing expressions and sample new candidates from this model.
This perspective connects our work to recent advances in EDA-GP, such as DAE-GP by Wittenberg et al. \cite{wittenberg2023denoising}, which uses a denoising autoencoder to model and resample promising programs. While we share the EDA principle, our implementation differs in using a VAE-based generative model with an explicit latent distribution and in incorporating domain knowledge through predicate-based filtering of candidate expressions. SEGVAE is algorithm, based on a variational autoencoder architecture with Long Short-Term Memory (LSTM) \cite{hochreiter1997long} as decoder and encoder.

Key SEGVAE aspects include:

(1) Noise Robustness: Unlike traditional symbolic regression models, SEGVAE demonstrates robustness against high noise levels, achieving performance levels comparable to DSO on standard benchmarks.

(2) Prior Knowledge Integration: SEGVAE allows users to encode domain-specific knowledge through predicate conditions, improving search efficiency and the relevance of generated expressions. This effectively narrows the search space, increasing the likelihood of finding meaningful symbolic representations.

(3) Pre-Training on Formula Structures: Pre-training enables the model to generate structurally valid formulas, facilitating faster and more accurate optimization for the target data without requiring extensive recalibration.

(4) Bank of Best Formulas (BBF): SEGVAE maintains a repository of optimal formulas, refining the variational autoencoder (VAE) over successive training epochs and allowing it to converge on the best-fit expressions.

We use a variational autoencoder for sequences in Polish notation (operators precede operands). The encoder and decoder are single-layer LSTMs with 64 hidden units; the latent code is 128-dimensional—a size that performed best in ablations while keeping training stable. We cap expression length at 30 tokens. SEGVAE takes as input a tokenized structure. Tokens represent operators, including basic arithmetic and functions like trigonometric and logarithmic operations, while variables and constants are also incorporated as tokens. These choices follow the original SEGVAE implementation and its ablation guidance, but are fixed here to our materials setting for reproducibility.

We used the same token list ['add', 'sub', 'mul’, 'div','sqrt','cos','exp','pow’] for SEGVAE train process and inference, where 'add' is '+', 'sub' is '-', 'mul’ is '\(\cdot\)', 'div' is '/', 'sqrt' is '\(\sqrt{}\)','cos' is kosinus function, 'exp' is an exponential function, 'pow’ is power operator. Token list is one of the hyperparameters to train SEGVAE; the rest we set by default, as in \cite{popov2023symbolic}. 

Since for scientific data, some prior knowledge on the functional dependency is usually available, the SEGVAE easily applies it to search for formulas. The SEGVAE framework distinguishes itself by incorporating domain knowledge directly through predicates, which act as structural filters within the training process. These predicates constrain the expression search space to reflect anticipated functional forms, aligning formula generation with known physical or functional constraints. We applied only general constraints on the final formula. In our case, (i) definedness on the working domain, (ii) no NaN/Inf, and (iii) physically reasonable behavior at large separations. These constraints sharply reduce the search space and improve convergence on our task. 

The training of SEGVAE involves two main steps: pre-training and the main training cycle. In pre-training, the model is exposed to random sequences of tokens to learn valid formula structures. The main training phase involves sampling candidate formulas, filtering them based on defined predicates, and evaluating them for accuracy on a target dataset using mean squared error (MSE). The best-performing formulas are stored in a “Bank of Best Formulas” and used to fine-tune the VAE, gradually focusing the model on generating high-fidelity expressions that fit the target data.

After training, SEGVAE generates candidate expressions that fit the target dataset, balancing between formula accuracy and complexity. A complexity metric is calculated for each candidate based on token types, helping to identify expressions that offer simplicity without sacrificing fit quality. SEGVAE outputs a Pareto front of formulas based on mean absolute error (MAE) and complexity available for used to choose the optimal function in terms of explainability and accuracy. Final expressions are selected from a Pareto frontier of accuracy versus complexity, allowing for informed selection based on the user's interpretability needs. In this work, the final equation is selected using a two-stage criterion:
(i) numerical selection: we first identify the simplest expression whose validation MAE is within 5\% of the best MAE on the Pareto front 
(ii) expert screening: among near-optimal expressions, we prefer formulas that satisfy additional physics-motivated sanity checks (e.g., definedness on the full domain, absence of NaN/Inf, and physically reasonable asymptotic behavior at large separations). When multiple candidates satisfy these criteria, we report the most compact one for readability. This selection principle (accuracy–complexity trade-off with preference for interpretable, physically meaningful forms) is common in physics-oriented symbolic regression; see, e.g., \cite{angelis2024reassessing}, where closed-form relations are chosen with emphasis on small complexity, interpretability, and reflecting underlying physical mechanisms.

One single structure of \ce{MoS2} 8x8 superlattice contains 192 atoms, thus in total 576 variables. 
To construct generalizable and interpretable formulas for formation energy per site and HOMO-LUMO gap, we do a similar trick of dimensionality reduction as in the sparse representation paper \cite{kazeev2023sparse}. As a general formula frame search for formation energy, we construct for pairwise interaction functions as an analogy to Coulomb interaction. In our setting (pairwise defect interactions in 2D), SEGVAE focuses the search on physically plausible kernels (decaying/oscillatory components) and improves data efficiency. 
\begin{equation} \label{eq1}
E_{formation} = \sum_{i\in{N}}E_i + \frac{1}{2}\sum_{i,j\in{N}}V_{i,j}(r)
\end{equation}

We define formation energy per site as in \cite{huang2023unveiling}.
\begin{equation} \label{eq2}
E_{\textrm{per site}} = \frac{E_{formation}}{N}
\end{equation}

Where N is a number of defects in the superlattice.

\begin{figure}[H]
	\noindent
	\centering
	\includegraphics[width=15.6cm]{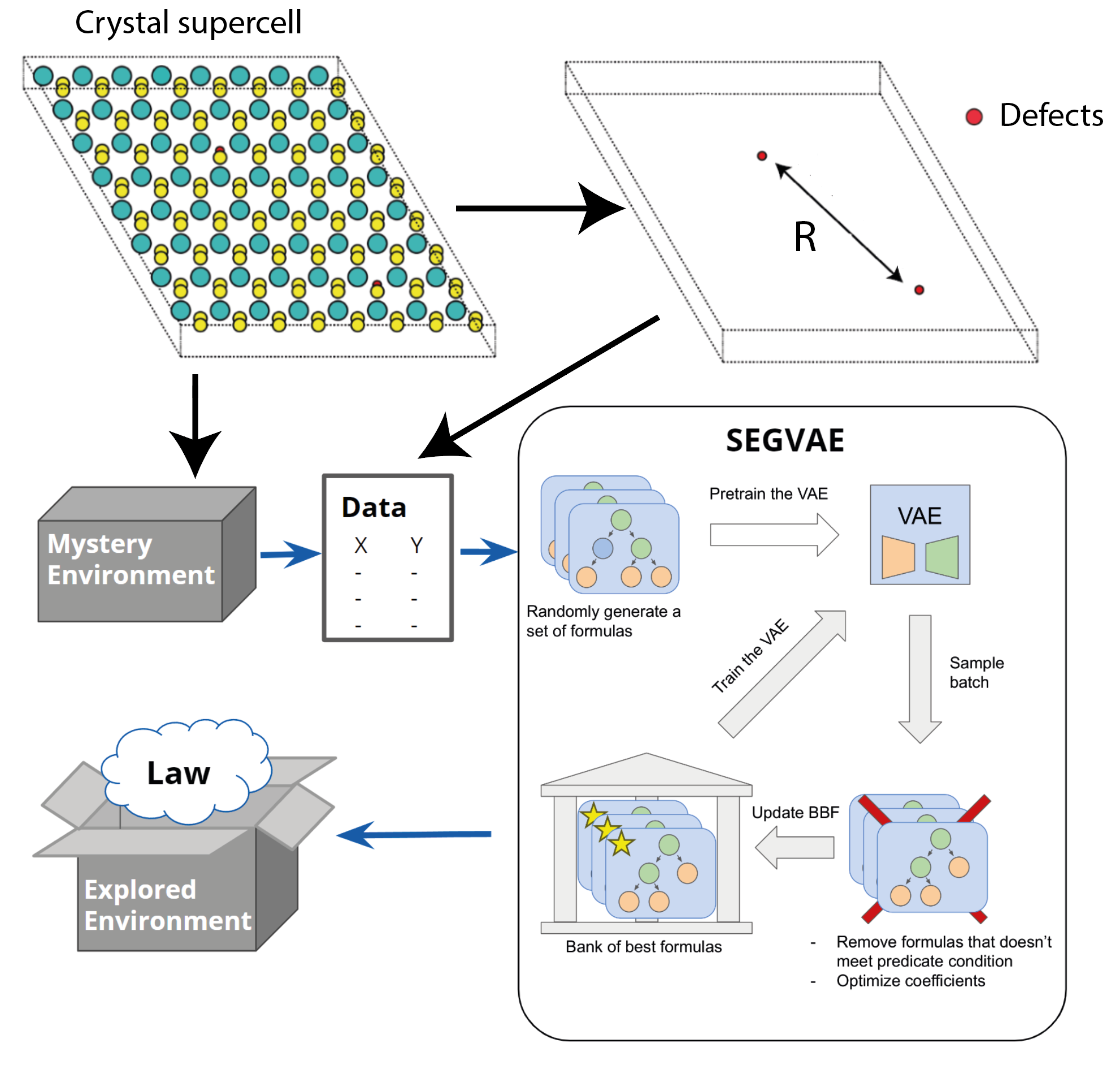}
	\caption{The SEGVAE application, architecture, and training scheme. Upper side: \ce{MoS2} with defects example.}
	\label{fig:SEGVAE_for materials}
\end{figure}

The frame formula for the HOMO-LUMO gap :

\begin{equation} \label{eq3}
E_{\textrm{HOMO-LUMO gap}} = \min_{i,j\in{N}} (E_{i,j}(r))
\end{equation}
We model the HOMO–LUMO gap using a “worst-case” (minimum) aggregation over pairwise contributions. This choice reflects that the global gap in a defective supercell can be governed by the most strongly gap-reducing defect interaction / localized defect state.
Thus, we postulate two main assumptions: We use Euclidean distance between the defects and consider interaction as a sum of pair defect interactions. For each defect interaction type, we select a small dataset with two defect structures of interest types. Figure \ref{fig:SEGVAE_for materials} illustrates schematically the interaction formula search for a simple interaction type. First, interested target values were calculated with DFT. The dataset for symbolic regression represents the distance between the defects at all possible positions, and as a target value, formation energy per site and HOMO-LUMO gap. 

\begin{equation} \label{eq4}
E_{formation} = const + V(r)
\end{equation}

For readability and interoperability, we report the simplest expression within 5\% of the best MAE on the Pareto front; in a few cases, this means not reporting the single best-MAE expression, while keeping the accuracy essentially unchanged. 
Here is an example of the formation energy per defect site as a function of Mo and S vacancy defects distance found by SEGVAE.

\begin{equation}
E_{\textrm{per site}}(x) = 4.895+(cos(0.5) \cdot  cos(8-2x) - 1) \cdot x^2  \cdot e^{-x}
\end{equation}

\begin{figure}[H]
	\noindent
	\centering
	\includegraphics[width=15.6cm]{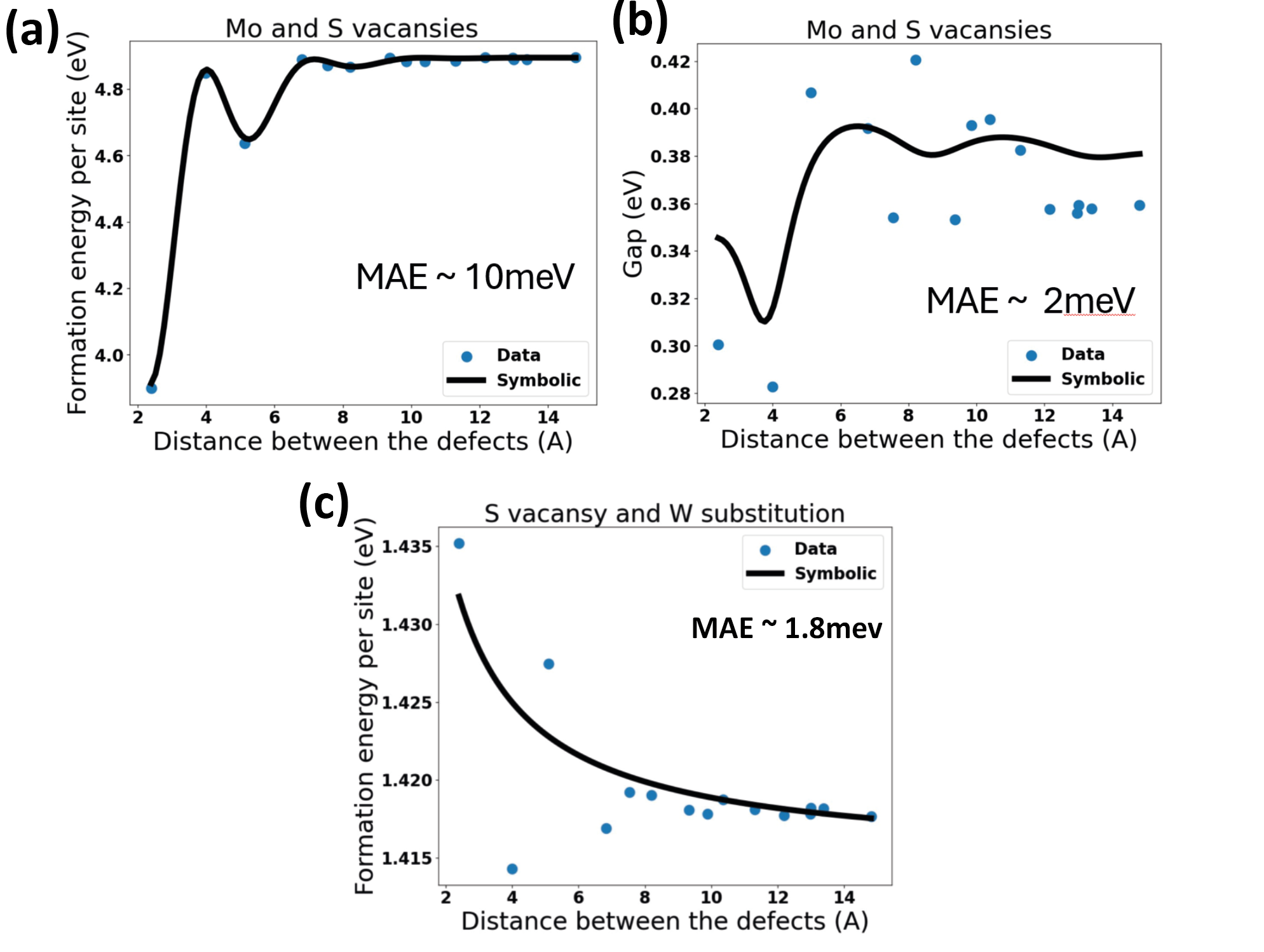}
	\caption{(a) Energy formation per site dependence on distance between Mo and S vacancies in \ce{MoS2} (b) Band Gap dependence from distance between Mo and S vacancies in \ce{MoS2} (c) Energy formation per site dependence from distance between W substitution and S vacancies in \ce{WSe2}. Data presented as circles. Note that panels (a–c) use different y-axis ranges (order-of-magnitude differences).}
	\label{fig:2Dmat}
\end{figure}

This formula shows how big the impact of vacancy interactions is on formation energy prediction, the graphical form with data points is illustrated in Figure \ref{fig:2Dmat}(a). 
%While a set of discrete energy values as a function of defect separation provides useful empirical insight, it is the discovery of an explicit analytical expression via symbolic regression that elevates this observation to the level of a physically interpretable model. The functional form captures not only the trend but also the oscillatory structure of defect interactions, reminiscent of known physical potentials. This transforms our numerical results into a hypothesis about the underlying physical law governing defect–defect coupling in 2D semiconductors.

%Interestingly, the symbolic regression yielded an analytic expression for the formation energy that includes an oscillatory term, multiplied by an exponentially decaying envelope. This structure is reminiscent of RKKY-type interactions and suggests the presence of quantum interference effects or Friedel-like oscillations \cite{friedel1952xiv} in the charge density redistribution around defects. Such behavior indicates that the defect–defect interaction is mediated not only through direct electrostatics but also via the electronic environment of the 2D lattice.

While a set of discrete DFT-computed values for defect formation energy as a function of separation provides empirical insights, the discovery of an explicit analytical expression via symbolic regression elevates the result to the level of a physically meaningful hypothesis. The resulting symbolic expression includes a product of an oscillatory term and an exponentially decaying envelope — a structure remind well well-known physical interaction potentials such as the Ruderman–Kittel–Kasuya–Yosida (RKKY) interaction \cite{ruderman1954indirect} and Friedel oscillations \cite{friedel1952xiv}.
The general form of RKKY interaction is: 
\begin{equation}
H(\mathbf{R}_{ij}) = \frac{\mathbf{I}_i \cdot \mathbf{I}_j}{4} \frac{|\Delta_{k_m k_m}|^2 m^*}{(2\pi)^3 R_{ij}^4 \hbar^2}
\left[
    2 k_m R_{ij} \cos(2 k_m R_{ij}) - \sin(2 k_m R_{ij})
\right],
\end{equation}
Where H represents the Hamiltonian, \ce{R_i_j} is the distance between the nuclei i and j, Ii is the nuclear spin of atom i, \(\Delta_{k_m k_m}\) is a matrix element that represents the strength of the hyperfine interaction, m* is the effective mass of the electrons in the crystal, and km is the Fermi momentum.

Although Friedel oscillations are formally defined as spatial modulations in the electron density \(\delta \rho(r) \propto \cos(2k_F r + \phi)/r^d\) induced by an impurity or defect, where \(k_F\) is Fermi wave vector, \(\phi\) is phase, r is the distance from the defect. In systems with interacting defects, the Friedel-type modulation of the electronic environment can lead to oscillatory variations in the formation energy as a function of defect separation.  These forms are characteristic of quantum interference and screening effects in many-body electron systems.
This similarity suggests that the interaction between defects in the 2D lattice is not purely electrostatic or local but is mediated through the delocalized electronic structure of the host material. Indeed, in monolayer transition metal dichalcogenides (TMDs) such as \ce{MoS_2} and \ce{WSe_2}, recent experimental and theoretical studies have demonstrated Friedel-like oscillations and long-range oscillatory strain fields around atomic defects \cite{lee2020deep} \cite{power2013indirect}, supporting the notion that quantum interference plays a role in defect–defect coupling.

In Figure \ref{fig:2Dmat}(b-c) we plot the pairwise contribution to the HOMO--LUMO gap as a function of the defect separation r together with the SEGVAE fit. For the HOMO–LUMO gap we obtain the following pairwise kernel (an example for the same interaction as in  Figure \ref{fig:2Dmat}(b). 

\begin{equation}
E_{\mathrm{gap}}(r)\approx 0.3601
-\frac{0.99}{5r + 10\cos\!\left(\frac{4}{3}r + 4\right)}
+\frac{r}{(r+4)^2 + 6(6-r)^2}.
\end{equation}

As a representative example of the discovered closed-form kernels, for the W substitution and S vacancy defects SEGVAE yields (Figure \ref{fig:2Dmat}(с)):

\begin{equation}
E_{\mathrm{gap}}(r)\approx 1.1611
+\frac{8}{\,r - 10^{\,6-r} - e^{6}}
+\frac{0.0485}{r}.
\end{equation}

For \ce{WSe_2} and other materials we follow the same procedure and repeat it for each interaction type. 

The different fit quality across Figure \ref{fig:2Dmat}(a–c) can be explained by both physics and the modeling assumptions. Vacancy–vacancy formation energy typically varies smoothly with separation and is well captured by a pairwise distance-based interaction kernel; the resulting decaying/oscillatory form is consistent with screened long-range electronic responses (RKKY/Friedel-type behavior) discussed above. In contrast, the HOMO–LUMO gap is more sensitive to defect-induced electronic states and their reordering/hybridization, which can lead to regime changes and clustered target values; therefore, a single smooth function of distance may not capture all variations equally well. Finally, substitution–vacancy interactions are affected by additional latent factors not explicitly modeled here (e.g., whether defects reside in the same or different atomic planes, local relaxations, and chemical asymmetry). Since we intentionally restrict descriptors to distance and defect types for interpretability, these latent degrees of freedom appear as irreducible scatter, lowering the apparent fit quality in Figure \ref{fig:2Dmat}(b) and (c). 

Therefore, the symbolic model derived here does not simply interpolate the data but reveals a functional form aligned with known physical principles, suggesting that symbolic regression can serve as a hypothesis generation mechanism — potentially uncovering new physical laws in systems where the governing interactions are not yet fully understood.

% распистаь как получалось HOMO-LUMO и какие там сложности?

\section{Results and comparison}

Each selected SEGVAE formula is relatively simple and interpretable. For simplicity in this study, we use only distance and atom types to construct the final function. However, accuracy still can be improved, for example in S (or Se) \ce{MoS2} (\ce{WSe2}) layer we did not distinguish between the upper or lower layers in the structure, but we know that defects interact differently if they are in one plane or in both. This can be taken into account by adding more interaction classes and separating cases where defects are on the same or different planes. Or in another way, perform additional coefficient optimization inside formulas.  
In total, for each interaction type, we had up to 20 structures to learn pairwise interaction laws in functional form. Thus, the total number of structures for \ce{MoS2} and \ce{WSe2} was less than 300 structures to learn all interactions presented in the dataset.

\begin{figure}[H]
	\noindent
	\centering
	\includegraphics[width=15.6cm]{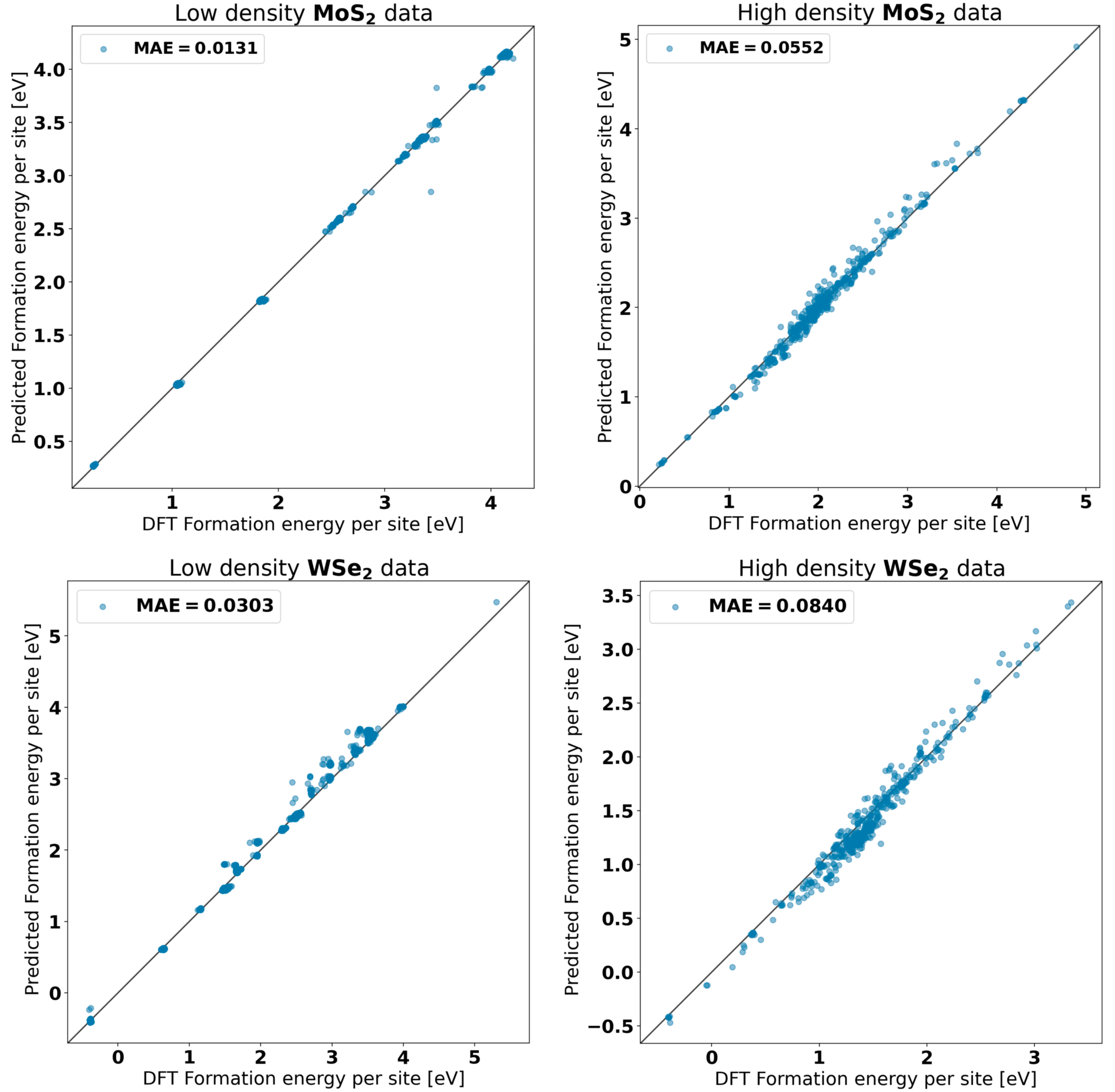}
	\caption{ Prediction results for energy formation per site for \ce{MoS2} and \ce{WSe2}. Left panels: Low-density data where a number of defects per structure is less or equal to 3. Right panels: Results on high-density defect structures where the number of defects is between 4 and 25.}
	\label{fig:Energy_formation}
\end{figure}

\begin{figure}[H]
	\noindent
	\centering
	\includegraphics[width=15.6cm]{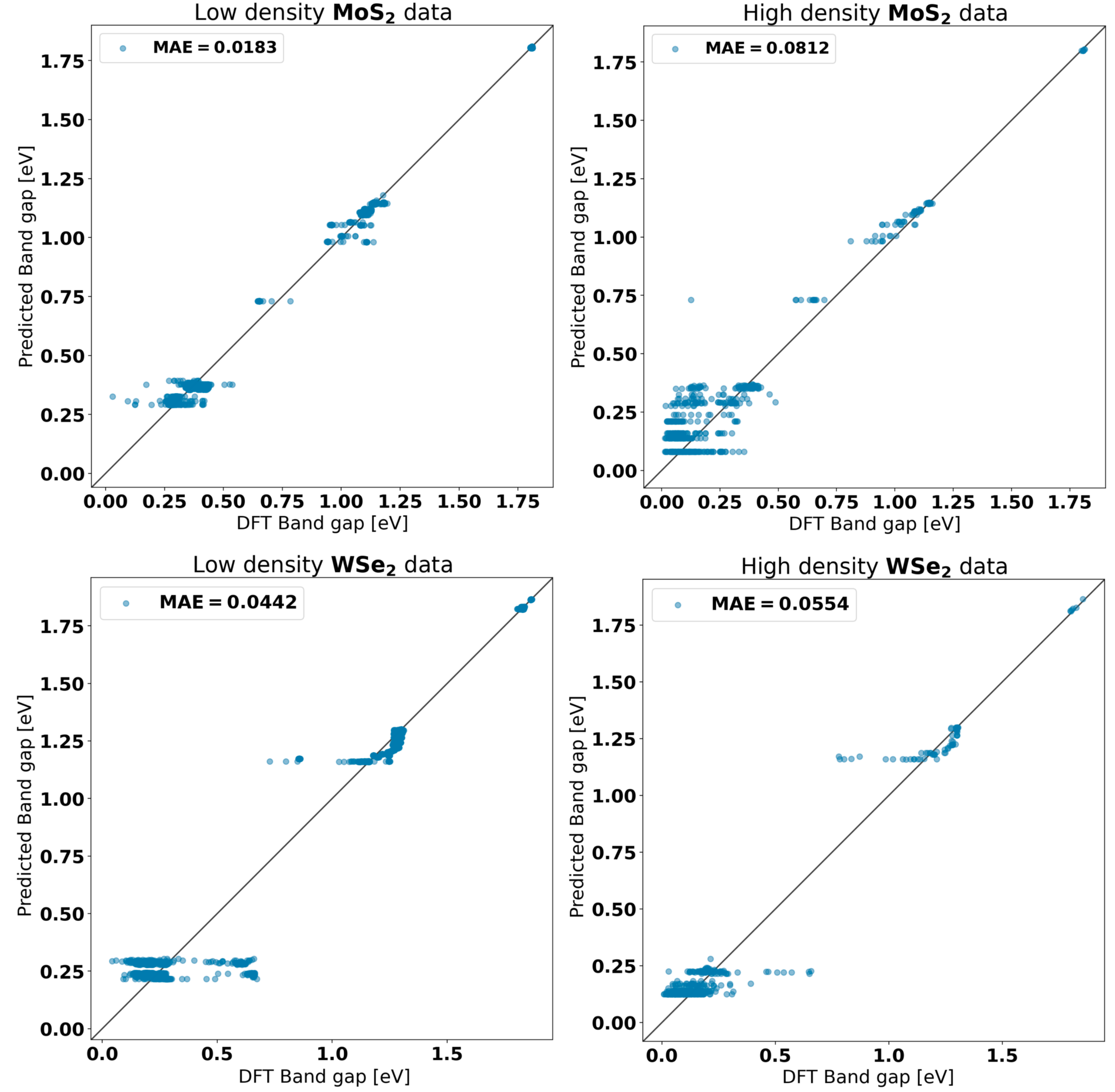}
	\caption{ Prediction results for Band Gap for \ce{MoS2} and \ce{WSe2}. Left panels: Low-density data where a number of defects per structure is less or equal to 3. Right panels: Results on high-density defect structures where the number of defects is between 4 and 25. }
	\label{fig:GAP}
\end{figure}

The general formula for formation energy per site obtained by substituting all the interaction laws found in Equation \ref{eq1} is general for any structure with any type of defects in the data set. Test results on high (more than 3 defects in supercell) and low density (less or equal to 3 defects in supercell) datasets, presented in Figure \ref{fig:Energy_formation}, right and left side correspondingly. Generalization on higher density defects data gives higher MAE, but still better results than all GNNs on full structure representation.
Figure \ref{fig:GAP} presents the results of our approach for HOMO-LUMO gap prediction based on Equation \ref{eq3}. We see that generalizing to data with a higher density of defects, the MAE error increased for both targets, and the same trend is true for GNNs. The increase in error is associated with the increasing influence of many-particle interactions with increasing defect density.
The clustered structure of band-gap values in Figure \ref{fig:GAP} can also reflect distinct electronic regimes associated with different defect species (e.g., vacancies vs substitutions and the chemical identity of the missing/substituting atom). Different defect types introduce defect states at different energies and with different localization, which can shift the gap into separate value ranges. Combined with the ‘min’ aggregation in Equation \ref{eq3}, this naturally leads to a multi-regime target and explains why a single smooth distance-only kernel may underfit band-gap trends compared to formation energies.

Table~\ref{tab:results-mae} summarizes the accuracy of the proposed pairwise-only symbolic model against GNN-based baselines on the same benchmark. For formation energy per site, the symbolic approach achieves errors comparable to MEGNet trained on sparse representations on the low-density subset, while remaining competitive in the high-density regime. For the HOMO--LUMO gap, the symbolic model performs competitively with SchNet, GemNet, and CatBoost across most settings, and is comparable to MEGNet on the full representation for several materials; the largest discrepancy is observed for \ce{WSe2} in the low-density case.

Importantly, our symbolic interactions are identified using only a small number of two-defect structures, after which the model is evaluated on low- and high-density defect configurations via aggregation. This highlights the ability of the proposed physically motivated construction to extrapolate from sparse pairwise data to more complex defect environments. Reported values for \textbf{Symbolic} are mean(std) over the 10 benchmark folds, matching the baselines’ evaluation protocol.

% проверить цифры в таблице

\begin{table}[H]
\centering
\captionsetup{font=small}
\footnotesize
\setlength{\tabcolsep}{6pt}
\renewcommand{\arraystretch}{1} 
\sisetup{detect-weight=true}
\begin{tabular}{cc|*{2}{S[table-format=4(3)]}
                S[table-format=3(2)]
                S[table-format=3.1(2.1)]
                S[table-format=3(2)]
                S[table-format=3(2)]}
\toprule
\multicolumn{8}{c}{Formation energy per site MAE, meV; lower is better} \\ \midrule {Material} & {Density} & {SchNet} & {GemNet} & {MEGNet} & {CatBoost} & { MEGNet(Sparse)} & \textbf{Symbolic} \\
\midrule

\ce{MoS2} & high & 321(100) & 535(206) & 136(22) & 136(5) & 23(5) & \textbf{\num{50(5.8)}} \\
\ce{WSe2} & high & 536(123) & 575(181) & 112(33) & 162(6) & 23(4) & \textbf{\num{74(9)}} \\
\ce{h-BN} & high & 1442(68) & 697(315) & 496(229) & 363(17) & 127(16) & \textbf{\num{295(38)}} \\
\ce{GaSe} & high & 245(12) & 230(41) & 107(25) & 103(4) & 48(7) & \textbf{\num{98 (7.3)}} \\
\ce{InSe} & high & 268(19) & 247(26) & 95(27) & 137(5) & 35(2) & \textbf{\num{70(7)}} \\

\hline
\ce{MoS2} & low & 65(5) & 44(14) & 58(11) & 12.6(0.4) & 4(1) & \textbf{\num{13(0.6)}}\\
\ce{WSe2} & low & 85(22) & 42(9) & 65(16) & 16.3(0.8) & 6(1) & \textbf{\num{30(1.3)}}\\
\bottomrule
\end{tabular}
\begin{tabular}{cc|*{2}{S[table-format=4(3)]}
                S[table-format=3(2)]
                S[table-format=3.1(2.1)]
                S[table-format=3(2)]
                S[table-format=3(2)]}
\toprule
\multicolumn{8}{c}{HOMO -- LUMO gap MAE, meV; lower is better} \\ \midrule
 {Material} & {Density} & {SchNet} & {GemNet} & {MEGNet} & {CatBoost} & {MEGNet(Sparse)} & \textbf{Symbolic} \\
\midrule

\ce{MoS2} & high & 204(121) & 174(111) & 54(4) & 71(4) & 39(4) & \textbf{\num{81(10)}}\\
\ce{WSe2} & high & 186(177) & 268(182) & 47(3) & 106(6) & 38(4) & \textbf{\num{55(6)}}\\
\ce{h-BN} & high & 244(24) & 227(6) & 233(4) & 208(3) & 260(14)& \textbf{\num{240(31)}} \\
\ce{GaSe} & high & 309(83) & 196(11) & 178(8) & 173(4) & 194(11) & \textbf{\num{201(18)}} \\
\ce{InSe} & high & 214(69) & 178(22) & 156(7) &  155(1) & 167(15) & \textbf{\num{170(14)}} \\
\hline
\ce{MoS2} & low & 187(180) & 46(42) & 30(2) & 26.7(0.8) & 5.7(0.2) &\textbf{\num{18 (0.5)}}\\
\ce{WSe2} & low & 236(224) & 64(46) & 32(5) & 18.3(0.6) & 8.1(0.6) & \textbf{\num{44(2)}}\\
\bottomrule
\end{tabular}

\caption{Accuracy comparison on the benchmark for formation energy per site and HOMO--LUMO gap. For our method Symbolic, we report MAE with standard deviation across 10-fold splits of the benchmark (MAE(std)). All baseline results are taken from \cite{kazeev2023sparse} as reported. Lower is better.}
\label{tab:results-mae}
\end{table}

To probe the impact of residual many-body effects, we additionally consider a lightweight three-defect correction on \ce{MoS2} in the low-density regime. Specifically, we keep the discovered symbolic pairwise expressions fixed and estimate a single additive intercept using the training portion of each fold (closed-form); we then report mean$\pm$std across folds. This post-hoc calibration does not require any additional symbolic searches. A systematic extension of this three-defect construction to all materials and density regimes is left for future work.

For structures containing exactly three defects $a,b,c$, we construct the pairwise-only prediction as
\[
\hat y_{\text{pair}} = f_{ab}(r_{ab}) + f_{ac}(r_{ac}) + f_{bc}(r_{bc}),
\]
where each $f(\cdot)$ is the symbolic interaction identified from two-defect data.
We then apply the simplest three-defect correction
\[
\hat y_{\text{three-defect}} = \hat y_{\text{pair}} + c,
\]
with a single intercept $c$ fitted on the training split of each fold.
The test MAE is computed on the corresponding held-out split, and we report mean$\pm$std across the 10 benchmark folds.

On this subset, the three-defect correction reduces the formation-energy MAE from \num{13(0.6)} to \num{4.3(0.9)}~meV. This brings the error close to the GNN baselines reported for \ce{MoS2} low density in Table~\ref{tab:results-mae}.

\section{Conclusion}

Symbolic regression is a fascinating method for deriving functional dependencies, allowing users to construct formulas that describe specific effects with defined physical or semantic constraints. To keep this usable in practice, we contain formula growth via predicate-guided grammars, dimensional/symmetry checks, and Pareto selection over error vs. complexity, with a final symbolic simplification/constant refitting; and we reduce reliance on user intuition through predefined operator libraries and filtering under the same predicates. With well-structured data, symbolic regression can yield interpretable results that outperform many methods, including state-of-the-art graph neural networks, as we demonstrated in our example. By using SEGVAE as a symbolic regressor and breaking down the formula for the dependency of crystal properties on defect positions into pairwise interactions, we achieved MAE metric results that surpass MEGNet and closely approach MEGNet's performance when applied to data in a sparse representation.

Also, our approach can be used for material design tasks by placing defects in certain positions to modify physical properties. This enables an inverse–defect–placement workflow: given a target window for formation energy or the HOMO–LUMO gap, we search defect configurations using the SR surrogate (milliseconds per query). We have demonstrated such a coupling of SR with a genetic optimizer in \cite{karlinski2024prediction} where we integrated the symbolic method into a genetic algorithm to generate defects in \ce{MoS_2}.

A key advantage of this approach lies in the interpretability of the result and computational speed, which can exceed that of GNNs by orders of magnitude and use less data consumption to learn functional dependencies. However, there are certain limitations. To accommodate new types of defects, one must derive formulas for new interactions, which grow quadratically with the number of defect types. Introducing new materials also requires crafting new formulas, as these are generally unique for each material. Moreover, the formulas themselves are not unique but are shaped by the subjective choices, intuition, and experience of the user. On the other hand, GNNs also require retraining, as they lack generalization capabilities for new materials and defects.

Our findings confirm the viability of deep symbolic regression methods for certain tasks in the natural sciences, showing that in some cases, the results can surpass established ML approaches.

\section{Acknowledgements}
The authors thank Pengru Huang from the National University of Singapore for running DFT simulations of new structures with two defects. 

\section{Funding}
The work was supported by the grant for research centers in the field of AI provided by the Ministry of Economic Development of the Russian Federation in accordance with the agreement 000000С313925P4E0002 and the agreement with HSE University № 139-15-2025-009

\section{Author contributions}
ML implemented SEGVAE, conducted computational experiments, interpreted the results, and wrote the manuscript. AU supervised the work. All authors contributed to the debate and analysis of the data and approved the final version.

\section{Data availability}

To fit pairwise interactions, we generated new structures that were not presented in the original dataset. We generated new structures h-BN, GaSe, InSe with all kinds of possible pairs defect configurations (see Table \ref{tab:defect-definitions}) on 8x8 supercell. Structures were relaxed using Vienna Ab initio Simulation Package (VASP) DFT simulation package \cite{kresse1996efficient}. The relaxation results are available in DOI:10.5281/zenodo.15806883.

\section*{Appendix}\label{Appendix}

\begin{figure}[H]
	\noindent
	\centering
	\includegraphics[width=15.6cm]{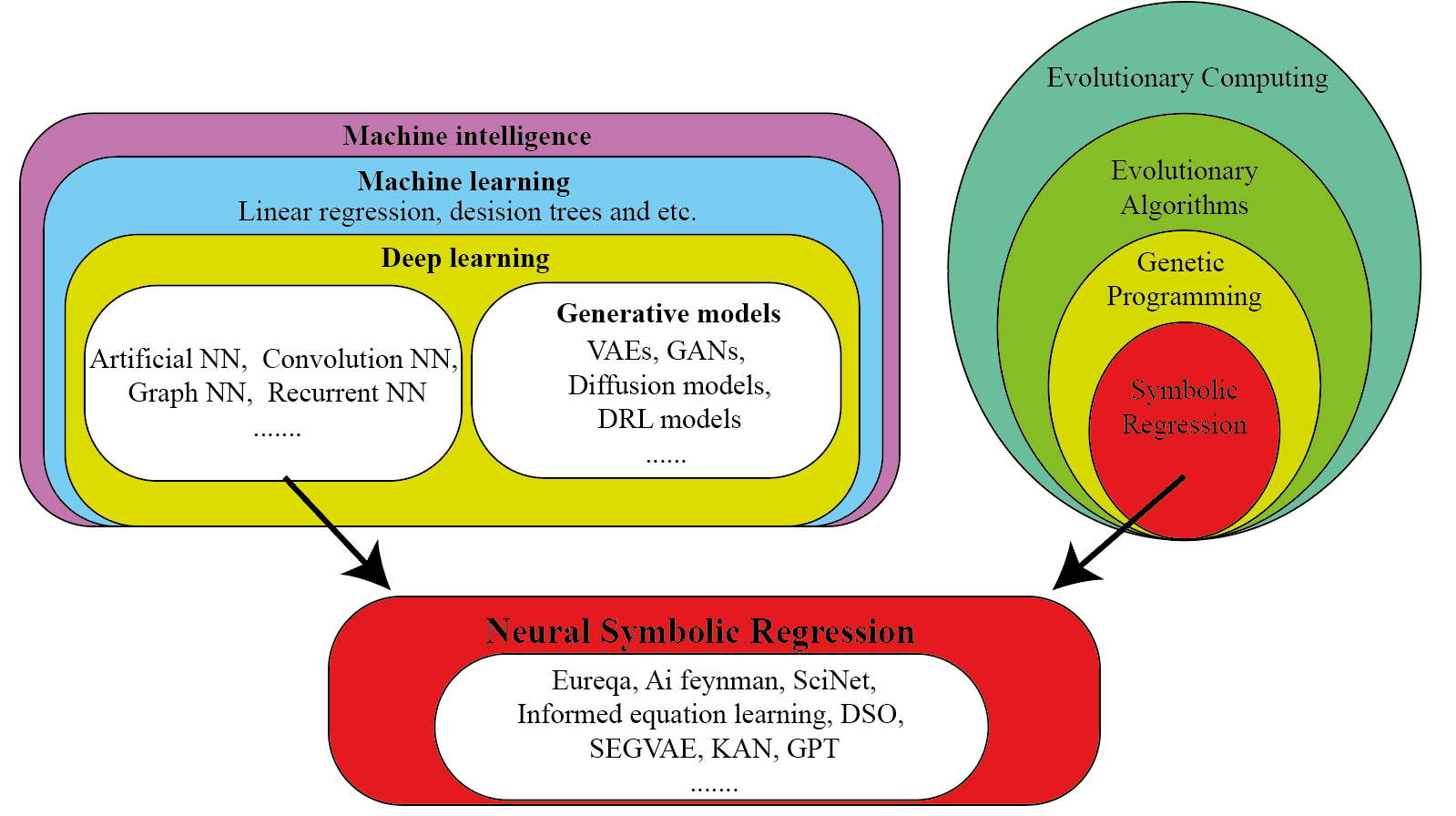}
	\caption{  ML and GP methods }
	\label{fig:Intro_ML}
\end{figure}

Variety of Symbolic regression algorithms is a reflection of the multitude of neural network types available; their adaptation for symbolic regression tasks offers a wide range of algorithmic options, each with its own advantages and limitations. This convergence of approaches is schematically represented in Figure \ref{fig:Intro_ML}.

Several symbolic regression approaches have demonstrated outstanding results, with one of the most well-known being the model for predicting physics formulas from Feynman's physics textbook \cite{AIFeynman}. This algorithm was later improved, resulting in \cite{AIFeynman2}. In \cite{udrescu2021symbolic} AIFeynman algorithm in combination with an ordinary differential equation solver was used to discover physical laws from video on toy examples.

PySR paper \cite{DBLP:journals/corr/abs-2006-11287} presents an approach that combines GNNs with symbolic regression to uncover explicit physical equations. By embedding strong inductive biases into GNNs and employing symbolic regression on the learned components, they demonstrated the ability to rediscover known physical laws and even derive a new analytic expression in a cosmology application. Under the hood, as a symbolic regressor, Eureqa was used. This work showcases how deep learning models, traditionally considered "black boxes," can be made interpretable, enhancing both generalization and the potential for new discoveries in fields ranging from physics to astrophysics. 

A series of papers \cite{DBLP:journals/corr/MartiusL16} and the latest version \cite{pmlr-v80-sahoo18a} were dedicated to an MLP-inspired approach. Informed Equation Learner (iEQL) that integrates domain-specific knowledge into symbolic regression models. Unlike traditional machine learning models that yield dense, uninterpretable representations, iEQL uses structured building blocks with a custom set of activation functions to derive compact, interpretable equations. This approach includes handling atomic functions with singularities (e.g., logarithm and division) and applying domain-dependent structured sparsity, allowing for more effective scaling to real-world scientific and engineering applications. Through experiments on simulated and real-world datasets, the iEQL demonstrated the ability to learn interpretable models with high predictive accuracy, especially in industrial applications like modeling power loss in electric machines and torque in combustion engines.

The variety of deep learning algorithms for various types of data spawns different SR approaches. 
One recent work \cite{li2022symbolic} treats the SR task as an image-to-expression translation problem. Instead of relying solely on searching mathematical spaces, Symbolic Expression Transformer SET represents sampled data as images and uses transformer-based encoding to convert visual patterns into symbolic expressions. SET introduces a large-scale dataset, carefully designed with non-overlapping training and test sets, but limited to two variables. 

Another advanced deep learning approach to symbolic regression is introduced in \cite{dsr}. Deep Symbolic Regression (DSR), which uses a reinforcement learning framework to recover exact mathematical expressions from data. Unlike traditional methods that focus on average performance, DSR applies a risk-seeking policy gradient to optimize for best-case outcomes, allowing it to identify highly accurate, interpretable models. This approach employs an autoregressive recurrent neural network (RNN) to construct expressions token by token, integrating domain-specific constraints in real time to avoid invalid formulations. By outperforming genetic programming and commercial tools like Eureqa on benchmark tasks, DSR demonstrates the potential of combining deep learning with symbolic methods to advance scientific discovery. DSR has recently been updated by introducing a genetic programming component \cite{mundhenk2021symbolic} (DSO).

Example of utilizing NLP methods in  \cite{biggio2021neural}, where transformer architecture was implemented, the authors propose to use pre-trained transformers \cite{NIPS2017_3f5ee243} to predict symbolic expression. This approach demonstrates comparable results with DSR, but compared with DSO on common benchmarks, the quality of the results is lower.

With the advent of large language models (LLMs), these too have been adapted for symbolic regression tasks. One of the pioneering works in this direction is \cite{valipour2021symbolicgpt}. Central to SymbolicGPT is its order-invariant embedding method, which uses a T-net \cite{qi2017pointnet} to convert datasets of varying sizes into robust representations, enabling the model to generate accurate expressions without relying on fixed data orders. Additionally, the model employs a generative pre-trained transformer to produce equation "skeletons" with constants optimized separately using the BFGS algorithm, streamlining computational load. The latest method \cite{li2024generative} was proposed last year, aiming to accelerate formula generation while achieving quality comparable to DSO. This was accomplished by using a GPT architecture to generate tokens, with features extracted using the SetTransformer \cite{lee2019set}, and a decoder that produces the sequence of the DSR training process. The authors demonstrated that their method could quickly generate formulas, and while its accuracy trails behind DSR and DSO, it outperforms SNIP \cite{meidani2023snip} and NeSymReS \cite{biggio2021neural} in precision.

However, a significant drawback of these GPT-based methods is the lack of accessible source code and model weights, preventing result validation and application to real-world problems.

\begin{figure}[H]
	\noindent
	\centering
	\includegraphics[width=15.6cm]{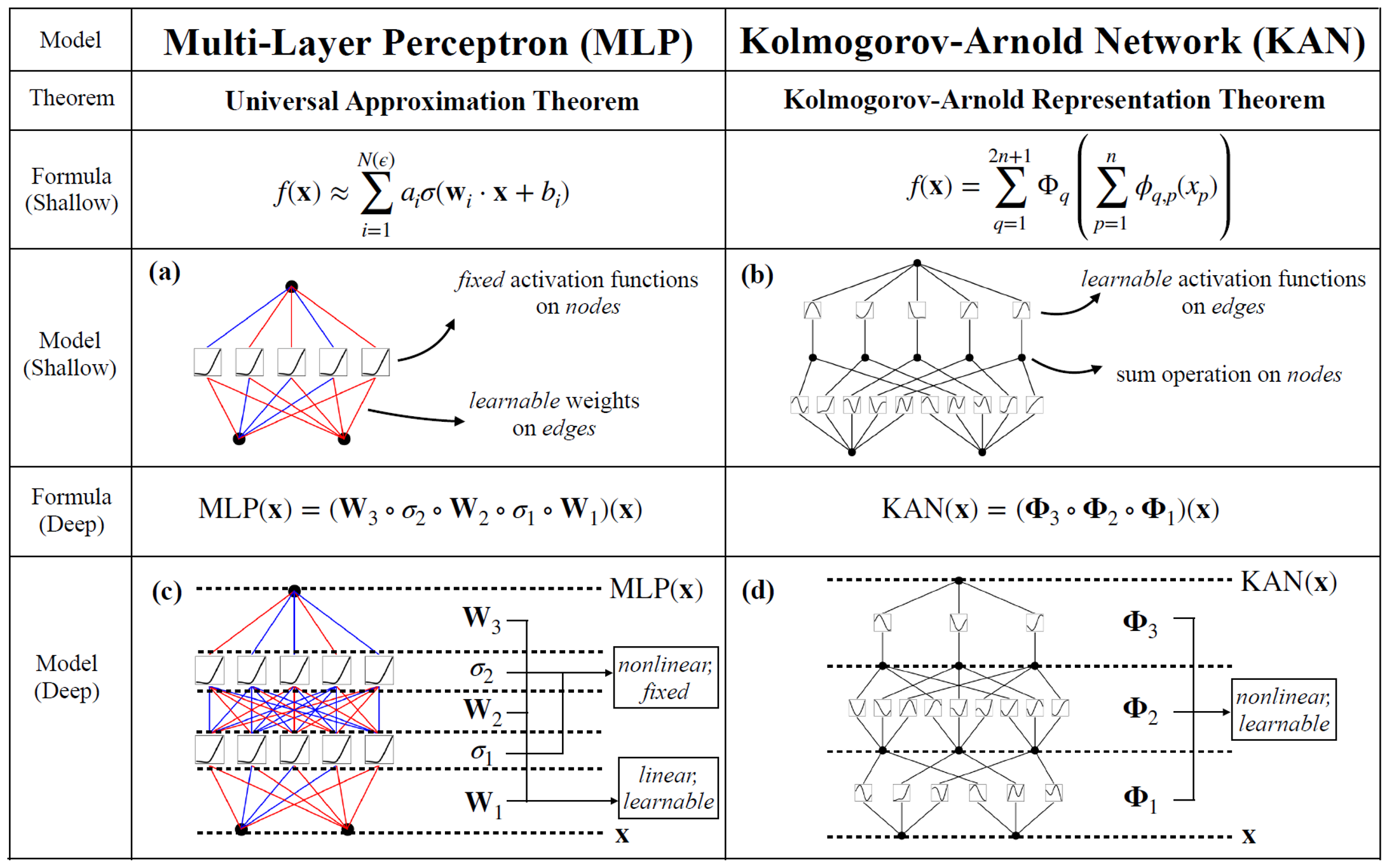}
	\caption{ MLP vs KAN. Taken from \cite{liu2024kan}}
	\label{fig:KAN}
\end{figure}

\begin{figure}[H]
	\noindent
	\centering
	\includegraphics[width=15.6cm]{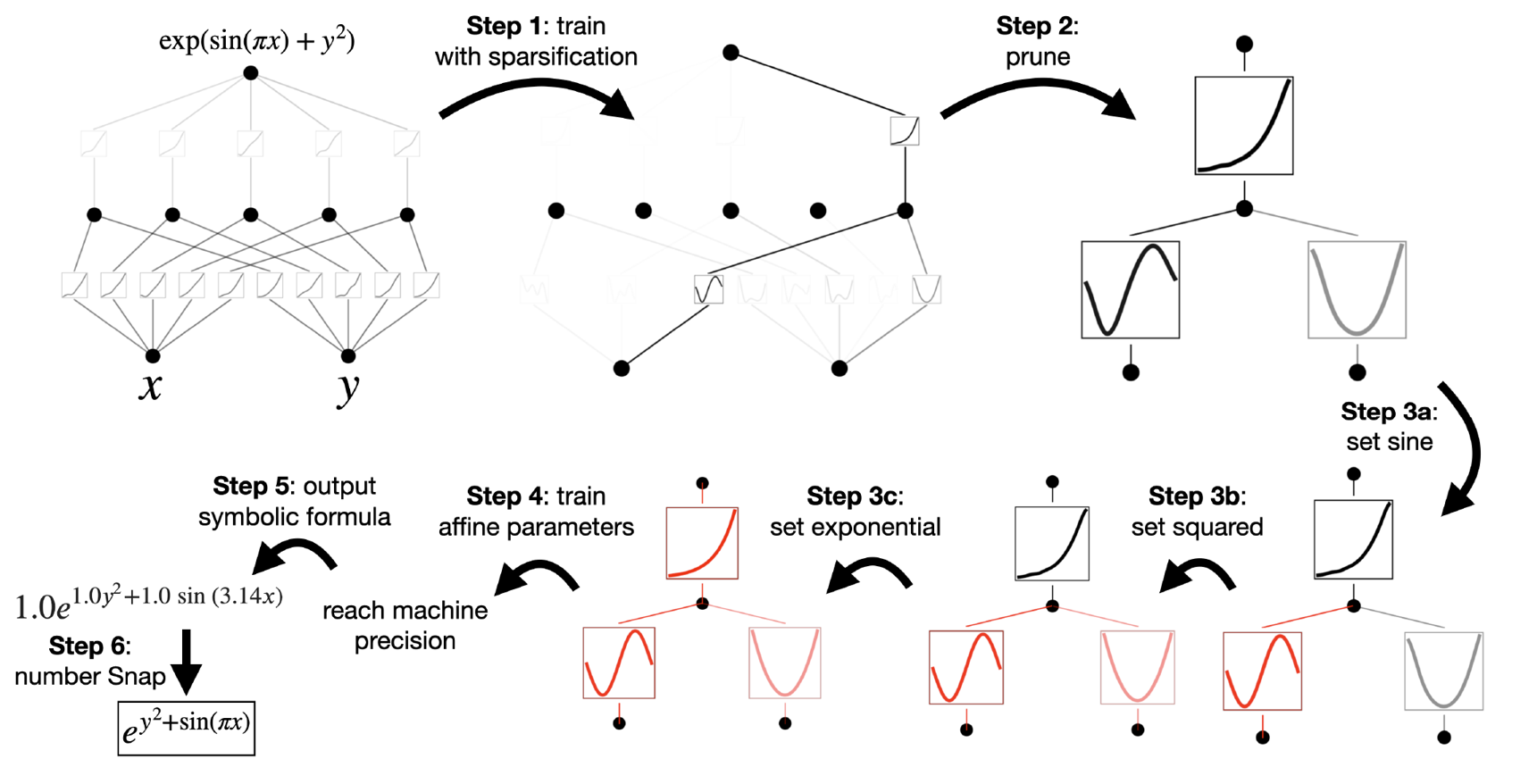}
	\caption{ KAN for symbolic regression. Taken from \cite{liu2024kan}}
	\label{fig:Symbolic_KAN}
\end{figure}

An elegant approach the worth mentioning is based on reimagining the MLP structure: Kolmogorov Arnold Networks (KAN), implemented in \cite{liu2024kan}. KANs redefine the traditional neural network structure by replacing fixed activation functions on nodes with learnable univariate functions on edges, modeled as splines. The difference is schematically shown in Figure \ref{fig:KAN}. This architecture reduces the curse of dimensionality by decomposing complex functions into simpler, one-dimensional components, allowing KANs to efficiently scale in high-dimensional settings. Unlike conventional MLPs, KANs are inherently interpretable, with symbolic simplification and regularization enabling clear visualization of learned functions, as shown in Figure \ref{fig:Symbolic_KAN}. Demonstrated in scientific tasks across mathematics and physics, KANs not only achieve high accuracy similar to MLPs but also offer insights into underlying laws and relationships.

In symbolic regression, the development of benchmarks that align with real-world scientific discovery is crucial to advancing the field. In \cite{matsubara2022rethinking}, address this need with Rethinking Symbolic Regression Datasets and Benchmarks for Scientific Discovery, introducing a new set of symbolic regression for scientific discovery (SRSD) datasets that incorporate physics-inspired formulas with realistic variable sampling ranges and added complexity through dummy variables. This approach reflects real-world scenarios, allowing models to demonstrate robustness and feature selection in noisy environments. To assess accuracy beyond traditional metrics, the authors propose Normalized Edit Distance (NED), which measures structural similarity between predicted and true equations, providing a more nuanced evaluation that correlates well with human judgment.

In parallel to the recent success of neural symbolic regression, genetic programming based symbolic regression has also seen noticeable progress in the last 3–5 years. First, several modern GP-SR frameworks were proposed with an emphasis on scalability, modularity, and practical usability. For example, Operon is a high-performance GP framework with efficient expression encoding and parallel offspring generation, and Bingo is a flexible GP-SR framework with swappable components (fitness, selection, constant optimization), designed to be easily adapted to different SR tasks \cite{burlacu2020operon,randall2022bingo}. Such toolchains make it easier to reproduce experiments and to compare GP baselines under consistent settings.

Second, the GP community went beyond the “classic” subtree crossover/mutation by developing stronger variation and tuning mechanisms. A representative example is GP-GOMEA, where mixing is guided by learned dependencies between parts of expressions; recent work also studied coefficient mutation inside GP-GOMEA to improve constant fitting and rediscovery of the underlying equations on SR benchmarks \cite{virgolin2022coefficient}. Another active direction is to guide GP search with semantics and constraints. For example, semantic schema-based GP proposes locality/semantic guidance for a more reliable search in symbolic regression \cite{zojaji2022semantic}. Moreover, formal constraint handling has been studied in counterexample-driven GP, where candidate expressions are automatically checked against user-defined constraints and violating examples are generated to refine evaluation and steer the search away from infeasible regions \cite{blkadek2022counterexample}. This direction is closely related to our predicate filtering (definedness checks, NaN/Inf rejection, and physically reasonable behavior), although we implement constraints as lightweight predicates compatible with a VAE-guided pipeline.

%Before presenting the results of SEGVAE approach in our case, we will cover the use of SR approach in natural science.

\printbibliography

\end{document}